\documentclass[sigconf]{acmart}

\usepackage{graphicx}
\usepackage{subfig}
\usepackage{multicol}
\usepackage{multirow}
\usepackage{threeparttable}
\usepackage{setspace}
\usepackage{float}
\usepackage{balance}

\AtBeginDocument{%
  \providecommand\BibTeX{{%
    \normalfont B\kern-0.5em{\scshape i\kern-0.25em b}\kern-0.8em\TeX}}}

\copyrightyear{2021}
\acmYear{2021}
\setcopyright{acmcopyright}\acmConference[MM '21]{Proceedings of the 29th ACM International Conference on Multimedia}{October 20--24, 2021}{Virtual Event, China}
\acmBooktitle{Proceedings of the 29th ACM International Conference on Multimedia (MM '21), October 20--24, 2021, Virtual Event, China}
\acmPrice{15.00}
\acmDOI{10.1145/3474085.3475643}
\acmISBN{978-1-4503-8651-7/21/10}

\begin{document}
\graphicspath{{figures/}}
\fancyhead{}
\title{MSO: Multi-Feature Space Joint Optimization Network for RGB-Infrared Person Re-Identification}





\author{Yajun Gao$^1$$^*$,\hspace{0.1em} Tengfei Liang$^1$$^*$,\hspace{0.1em} Yi Jin$^1$$^{\dag}$,
\hspace{0.1em} Xiaoyan Gu$^2$,\hspace{0.1em} Wu Liu$^3$,\hspace{0.1em} Yidong Li$^1$,
\hspace{0.1em} Congyan Lang$^1$}
\affiliation{%
 \vspace{0.1em}\institution{$^1$School of Computer and Information Technology, Beijing Jiaotong University \state{Beijing} \country{China}}
}
\affiliation{%
 \vspace{0.1em}\institution{$^2$Institute of Information Engineering, Chinese Academy of Sciences \state{Beijing} \country{China}}
}
\affiliation{%
 \vspace{0.1em}\institution{$^3$JD AI Research \state{Beijing} \country{China}}
}
\email{%
  {yajun.gao, tengfei.liang, yjin, ydli, cylang}@bjtu.edu.cn, 
  guxiaoyan@iie.ac.cn, 
  liuwu1@jd.com
}
\thanks{$^*$Both authors contributed equally to this research.}
\thanks{$^{\dag}$Corresponding author}

\renewcommand{\shortauthors}{Gao and Liang, et al.}

\begin{abstract}
  The RGB-infrared cross-modality person re-identification (ReID) task aims to recognize the images of the same identity between the visible modality and the infrared modality. 
  Existing methods mainly use a two-stream architecture to eliminate the discrepancy between the two modalities in the final common feature space, which ignore the single space of each modality in the shallow layers. 
  To solve it, in this paper, we present a novel multi-feature space joint optimization (MSO) network, which can learn modality-sharable features in both the single-modality space and the common space. 
  Firstly, based on the observation that edge information is modality-invariant, we propose an edge features enhancement module to enhance the modality-sharable features in each single-modality space. 
  Specifically, we design a perceptual edge features (PEF) loss after the edge fusion strategy analysis. 
  According to our knowledge, this is the first work that proposes explicit optimization in the single-modality feature space on cross-modality ReID task. 
  Moreover, to increase the difference between cross-modality distance and class distance, we introduce a novel cross-modality contrastive-center (CMCC) loss into the modality-joint constraints in the common feature space. 
  The PEF loss and CMCC loss jointly optimize the model in an end-to-end manner, which markedly improves the network's performance. 
  Extensive experiments demonstrate that the proposed model significantly outperforms state-of-the-art methods on both the SYSU-MM01 and RegDB datasets.
\end{abstract}




\maketitle


\section{Introduction}

RGB-infrared cross-modality person re-identification(ReID) aims to recognize the images of the same identity between the two modalities, the visible (RGB) modality and the infrared (IR) modality. 
With the widespread use of near-infrared cameras, it has attracted increasing attention and has great application value in the nighttime surveillance field. 
Due to the huge differences between two modalities (e.g., color, texture), how to reduce the modality discrepancy of the same identity is essential for the solution of cross-modality person ReID task. 

Existing methods mainly include two solving strategies. 
The first strategy uses a two-stream architecture to eliminate the discrepancy between the two modalities.  
Modality-sharable features are extracted via the weight-specific network, 
and then discriminative features for matching are embeded into the same feature space via the weight-shared network~\cite{HCML,dc,bdcc,Sphere,EDFL}. 
However, these methods mainly focus on feature learning in the final common feature space while ignoring the single space of different modalities. 
There is less work focus on the direct enhancement of modality-sharable features in each single-modality feature space. 
Moreover, although these methods focus on narrowing the distance of the same identity between two modalities in the feature space, they ignore the feature distance among different identities, 
making it difficult to distinguish different identities. 
There is less work focus on the efficient constraint of overall feature distribution. 
The second strategy extracts the modality-specific features and transforms them from one modality to another by a generator or a decoder
~\cite{cmGAN,AlignGAN,DRL,Hi-CMD,JSIA}. 
However, in the RGB-IR person ReID task, objects with the same color in the RGB image have different appearances in the IR image. 
Lack of stable mapping relationship and unavoidable noise cause unreliable image generation. 
Besides, this kind of methods usually has very slow convergence.

\begin{figure}[t]
  \centering
  \includegraphics[width=\linewidth]{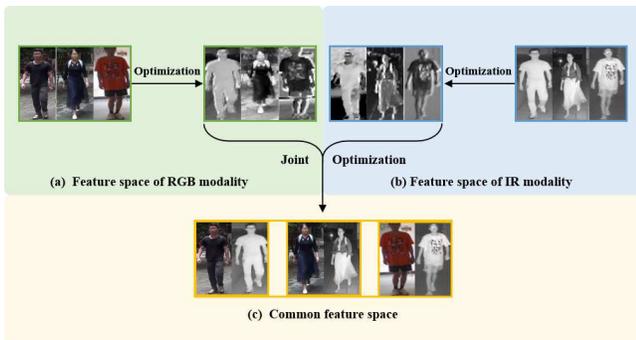}
  \caption{A high-level overview of our approach: 
  MSO network optimizes both single-modality feature space and common feature space in an end-to-end manner simultaneously. }
  \label{overview}
\end{figure}

In light of the above observations, we design a multi-feature space joint optimization (MSO) network that mines more modality-sharable features in each single-modality feature space and learns discriminative features without modality transfer. 
For the RGB-IR cross-modality task, information such as color can cause a larger gap, 
while the outline information of body and hair is an important focus in this task because it will not change between different modalities (i.e., modality-invariant). 
Some works focus on how to eliminate the negative influence of color information. 
In our work, considering that edge features can effectively describe the outline information, 
we design an edge features enhancement module. 
We introduce this module into the weight-specific network layer, so the ability of the modality-sharable feature extraction is improved in each single-modality feature space. 
Moreover, we analyze five edge fusion strategies in section~\ref{Analysis} and select the perceptual edge features (PEF) loss as the final enhancement method. 
Our PEF loss allows features to focus on more edge information while retaining other useful information by perceptual losses~\cite{perceptual}. 
After the weight-shared network layer, we introduce a novel cross-modality contrastive-center (CMCC) loss into the modality-joint constraints to increase the difference between cross-modality distance and class distance. 
Thus, in the common feature space, the features from different identities are more distinguishable while keeping features compact under the same identity. 
With the above two modules, the MSO model optimizes both the feature space of RGB and IR modality and the common space simultaneously in an end-to-end manner as shown in Figure~\ref{overview}. 

Through comparison with state-of-the-art methods, the proposed model shows excellent performance. 
In conclusion, the major contributions of our paper can be summarized as follows: 
\begin{itemize}
  \item We propose a novel multi-feature space joint optimization network to optimize both single-modality feature space and common feature space for effective RGB-IR cross-modality person ReID. 
  Extensive experiments prove the proposed model significantly outperforms state-of-the-art methods. 

  \item The perceptual edge features loss is proposed as an edge features enhancement module to preserve edge information of each modality.  
  To our knowledge, this is the first work that proposes an explicit optimization in the single-modality feature space on cross-modality person ReID task. 

  \item The cross-modality contrastive-center loss is introduced into the modality-joint constraints to learn a more suitable distribution in feature space, which has compact intra-class distribution and sparse inter-class distribution. 

\end{itemize}

\begin{figure*}[t]
  \centering
  \includegraphics[width=\linewidth]{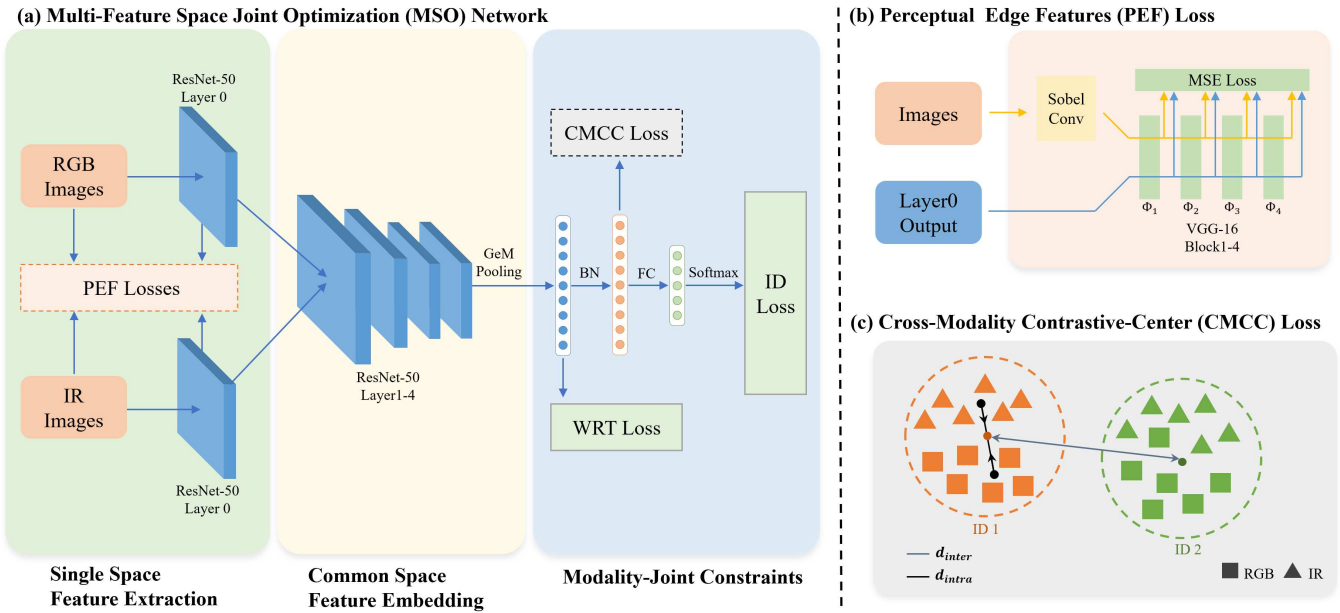}
  \caption{The framework of our proposed network: 
  (a) The overall structure of multi-feature space joint optimization network, 
  which consists of single space feature extraction, common space feature embedding, and modality-joint constraints. 
  (b) The flowchart of perceptual edge features loss. 
  (c) The diagram of cross-modality contrastive-center loss. 
  Different identities are represented by different colors.  
  $d_{intra}$ is the distance between the feature centers of different modalities of the same identity. 
  $d_{inter}$ is the distance between the feature centers of different identities. }
  \label{Whole_Model}
\end{figure*}

\section{Related Work}
\subsection{RGB-IR Cross-Modality Person ReID}

RGB-IR cross-modality person ReID aims to match person images captured by the visible cameras and the infrared cameras. 
Compared with RGB-RGB single-modality person ReID that only deals with visible images, the challenge of this task is how to bridge the huge gap between two different modalities. 
Existing work addresses the problem from different perspectives. 
Some early work focused on resolving channel mismatches between RGB images and IR images because RGB images have three channels (i.e., red, green, and blue). In contrast, IR images have only one channel. 
Wu et al.~\cite{Wu_2017_ICCV} proposed a deep zero-padding method to a uniform number of channels. 
Kang et al.~\cite{Kang} transformed RGB images into gray images and then proposed three different combinations to combine gray images and IR images as a single input. 
However, these hand-designed methods did not work well.
Then researchers repeat the single channel of IR images three times as input and pay more attention to discriminative feature learning. 

Nowadays, most methods use a two-stream architecture to eliminate the discrepancy between the two modalities. 
These methods learned modality-specific features through the shallow network layers that do not share weights 
and extracted modality-sharable features through the deep layers with weights sharing.
Ye et al.~\cite{HCML} learned multi-modality shareable feature representations with the two-stream CNN network, 
while Liu et al.~\cite{EDFL} fused the mid-level features from middle layers with the final feature of backbone. 
The MAC method~\cite{MAC} is proposed on top of a two-stream network to capture the modality-specific information. 
However, these methods only focus on feature learning in the final common feature space. 
Our model can optimize both single-modality feature space, and common feature space with an edge features enhancement module and the modality-joint constraints. 

Other researchers use the idea of Generative Adversarial Network (GAN)~\cite{GAN, CGAN, WGAN, SPGAN, PIX, CycleGAN} to transform modality. 
Hi-CMD model could extract pose-invariant and illumination-invariant features~\cite{Hi-CMD}. 
Dai et al.~\cite{cmGAN} generated modality-sharable representation through a minimax game with a generator and a discriminator. 
A dual-level discrepancy reduction learning scheme is proposed to project inputs from image space to the feature space~\cite{DRL}. 
Wang et al.~\cite{AlignGAN} generated fake IR images compared with real IR images through pixel alignment and feature alignment jointly. 
They also generated cross-modality paired-images with an instance-level alignment model and reduced the gap between different modalities~\cite{tsGAN}. 
However, the models with GAN introduce unavoidable noise and make the fake image generation unreliable. 

Considering that there exists correlations between two different modalities, Zhang et al.~\cite{DSCSN} obtained contrastive features by their proposed contrastive correlation. 
Feng et al.~\cite{DGD} imposed the cross-modality Euclidean constraint and identity loss to generate modality-invariant features. 
Lu et al.~\cite{ssft} decoupled features through a sharable-specific feature transfer network and then fused them with GCN~\cite{GCN}. 
Kansal et al.~\cite{SDL} learned spectrum-related information and obtained spectrum-disentangled representation. 
Wei et al.~\cite{CoAL} proposed an attention-based approach with multiple feature fusion. 
This kind of sharable-specific feature transfer and fusion methods needs additional modality information during testing to ensure performance. 
Li et al.~\cite{X} introduced a third X modality to feature space. 
Ye et al.~\cite{HAT} generated a third auxiliary grayscale modality from the homogeneous visible images. 
Ling et al.~\cite{CMM+CML} used class-aware modality mix to generate mixed samples for reducing the modality gap in pixel-level. 
Considering the similarities among gallery samples of RGB modality, Jia et al.~\cite{SIM} proposed the similarity inference metric as a re-ranking method, thus obtained relative high performance in the multi-shot setting.

\subsection{Loss in Cross-Modality Person ReID} 

The loss mostly used for cross-modality person ReID is identity loss~\cite{Zheng_2017_CVPR}, 
which treats ReID as an identity classification issue in training~\cite{GLTDAPR,BaoLCZM21,GanYG16}. 
Contrastive loss~\cite{contrastive} and triplet loss~\cite{triplet,Triplet0,Triplet1,Triplet2} focus on the distance between image pair. 
Some loss functions are built in terms of tasks~\cite{dc,bdcc,GanYG16,WangLLLLM20,GanZCC019}. 
Hao et al.~\cite{Sphere} used hypersphere manifold embedding with sphere loss~\cite{SphereFace,SphereReID,SphereL2} for metric learning. 
Similar to triplet loss, HP loss~\cite{HP} is proposed to minimize the distance between the cross-modality positive pair and maximize the distance between the cross-modality negative pair. 
Hetero-center (HC) loss~\cite{HC} is proposed to improve the intra-class cross-modality similarity without considering inter-class cross-modality similarity. 
Ling et al.~\cite{CMM+CML} optimized the network with identity classification loss, KL-divergence loss, and center-guided metric learning loss. 
The research results of these losses show that the best performance is to constrain the samples and the whole sample sets simultaneously. 
Inspired by the above observations, we introduce a novel cross-modality contrastive-center loss to form a modality-joint constraints section 
with identity loss and weighted regularization triplet loss~\cite{AGW} for RGB-IR cross-modality person ReID.
With the cross-modality contrastive-center loss, the features from different identities are more distinguishable while keeping features compact under the same identity. 

\section{The Proposed Method} \label{methodology}
In this part, we introduce the proposed multi-feature space joint optimization (MSO) network in detail.
Firstly, we describe the network structure in subsection~\ref{methodology_section_1}, and explain the feature extraction process and the location of two proposed modules. 
Then we introduce these two novel proposed losses, the perceptual edge features (PEF) loss (subsection~\ref{methodology_section_2}) and the cross-modality contrastive-center (CMCC) loss (subsection~\ref{methodology_section_3}), which correspond to the optimization of single and common feature space respectively.
In the last subsection~\ref{methodology_section_4}, we show and explain the complete formula of loss function during our training process.

\subsection{Overall Model Structure} \label{methodology_section_1}
The method of our paper mainly focuses on losses-based multi-feature space joint optimization. 
On the model side, we adopt the structural design of the previous method, the two-stream AGW ReID model~\cite{AGW}. 
As shown in Figure~\ref{Whole_Model}a, the two-stream structure is integrated into our MSO network with ResNet-50 as the backbone. 
And we divide the proposed network into three parts: single space feature extraction, common space feature embedding, and modality-joint constraints as illustrated in Figure~\ref{Whole_Model}a.

Let $rgb$, $ir$ denote the RGB modality and IR modality respectively. 
Let $X^m=\{x^m|x^m \in \mathbb{R}^{H \times W \times 3}\}$ denotes the RGB image set and the IR image set, 
where $m \in \{rgb, ir\}$, $H$ and $W$ are the height and the width of images. 
Each IR image also contains three channels by repeating its single channel three times as input. 
Suppose there are $B$ images in one batch during training. 
$x^m_i$ represents the $i$th image in an input batch, where $i \in \{1,2,...,B\}$. 
During training, each $x^{rgb}_i$ or $x^{ir}_i$ flows into its corresponding branch in the part of single space feature extraction (Figure~\ref{Whole_Model}a). 
Let $\mathcal{F}_{rgb}$ and $\mathcal{F}_{ir}$ denote the unshared layer0 of ResNet-50, the shallow layer features of each modality is extracted and defined as follows: 
\begin{equation}
  f^{rgb}_i = \mathcal{F}_{rgb}(x^{rgb}_i) 
  \qquad
  f^{ir}_i = \mathcal{F}_{ir}(x^{ir}_i)
  \label{F}
\end{equation}
$f^{m}_i$ denotes the modality-specific features, where $m \in \{rgb, ir\}$. 
Our proposed perceptual edge features (PEF) loss is designed to enhance the extracted features in single space. 

After that, $f^{rgb}_i$ and $f^{ir}_i$ are sent to the part of common space feature embedding. 
During common space feature embedding, we utilize the layer1 to layer4 of ResNet-50 to obtain modality-sharable features. 
Unlike the previous layer0 used separately for each modality, the feature embedding uses a shared structure design to extract modality-sharable features. 
Following ~\cite{AGW}, the non-local attention blocks ~\cite{Non-local} are inserted in the same positions as~\cite{AGW}. 
After that, we let the feature matrix through a generalized-mean (GeM) pooling layer~\cite{AGW}. 
For each $f^m_i$, we get a feature vector of 2048 dimensions. 
In modality-joint constraints, these 2048-dimensions vectors pass through BN layer~\cite{BN}, FC layer, and Softmax operation in turn. 
The Identity (ID) loss~\cite{Zheng_2017_CVPR}, weighted regularization triplet (WRT) loss~\cite{AGW}, and our designed cross-modality contrastive-center (CMCC) loss are calculated on these vectors. 
During testing, we sent query and gallery images into its corresponding network branch, respectively. 
Following ~\cite{AGW}, we obtain the features after the BN layer for calculating similarities of query and gallery.

\subsection{Perceptual Edge Features Loss} \label{methodology_section_2}

Considering that modality-specific information brings negative effects for RGB-IR person ReID task and is not easy to eliminate, 
we hope to find prior information that can be used to compact the RGB and IR modality space. 
Due to the modality invariance of outline information, which can be described by edge features, we use edge information as a self-supervised learning guide for feature enhancement. 
We propose perceptual edge features (PEF) loss to enhance the modality-sharable features and introduce PEF loss into single space feature extraction as shown in Figure~\ref{Whole_Model}. 

\begin{figure}[t]
  \centering
  \includegraphics[width=0.9\linewidth]{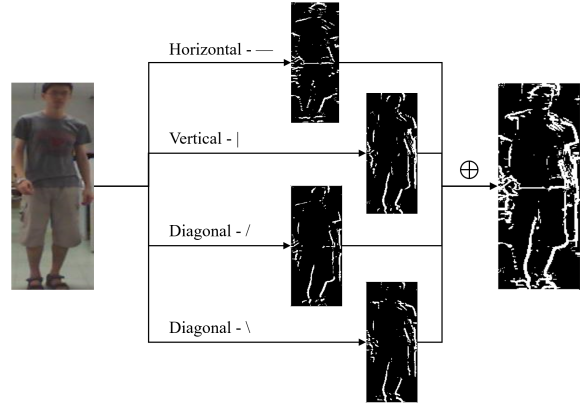}
  \caption{Sobel Convolution Module.}
  \label{sobel}
\end{figure}

PEF loss constrains features $f^{rgb}_i$ and $f^{ir}_i$ obtained by the unshared layer0 of ResNet-50. 
Specifically, we obtain edge features by the sobel convolution module for the enhancement of the modality-sharable features as shown in Figure~\ref{Whole_Model}b. 
As shown in Figure~\ref{sobel}, the sobel convolution module uses four classical sobel operators~\cite{Sobel} as convolution kernels. 
For each image, input it and then output a result containing four channels of edge features in different directions. 
The final edge features are obtained by adding these four channels together. 
We measure perceptual differences between shallow features $f^{m}_i$ and edge features for each image. 
As shown in the Figure~\ref{Whole_Model}b, the perceptual losses are computed by the block1 to block4 of the VGG-16 network~\cite{VGG}. 
Moreover, the VGG-16 network is pretrained on the ImageNet dataset and keeps weights non-learnable during the training of our proposed MSO model. 
Let $\phi = \{\phi_1, \phi_2, \phi_3, \phi_4\}$ represent the loss network as shown in the Figure~\ref{Whole_Model}b, 
$\phi_t (z)$ represents the feature maps of input $z$ with shape $C_t \times H_t \times W_t$, which is obtained by $t$th block of the network. 
Let $e_i^{rgb}$ and $e_i^{ir}$ represent the edge features extracted by sobel convolution module. 
$\mathcal{L}_{pef}$ can be formulated by: 
\begin{equation}
  \mathcal{L}_{pef} = \sum_{t=1}^4 \ell_{pef}^{\phi_t}(f_i^{rgb}, e_i^{rgb}) + \sum_{t=1}^4 \ell_{pef}^{\phi_t}(f_i^{ir}, e_i^{ir})
  \label{pef}
\end{equation}%

When inputting $f_i^{m}$ and $e_i^{m}$ into the loss network, $\ell_{pef}^{\phi_t}(f_i^{m}, e_i^{m})$ is calculated by mean square error (MSE) loss function: 
\begin{equation}
  \ell_{pef}^{\phi_t}(f_i^{m}, e_i^{m}) = \frac{1}{C_t H_t W_t} \sum_{c=1}^{C_t} \Vert \phi_t^c (f_i^{m}) - \phi_t^c (e_i^{m}) \Vert_F^2 
  \label{l}
\end{equation}%

During training, with the convergence of $\mathcal{L}_{pef}$, 
the feature maps $f_i^{m}$ of the unshared layer0 are encouraged to be similar to the edge information at the perception level. 
After that, the outline information can be directly enhanced. 
Thus, the trained network can extract more modality-sharable features in the feature space of each modality by PEF loss. 


\subsection{Cross-Modality Contrastive-Center Loss} \label{methodology_section_3}
Although the difficulty of cross-modality tasks lies in narrowing the distance between features in different modalities, 
we should also pay the interval between features of different identities special attention. Otherwise, it will reduce the accuracy of recognition. 
Inspired by the contrastive loss~\cite{contrastive}, we introduce a novel cross-modality contrastive-center (CMCC) loss into the modality-joint constraints. 
CMCC loss can avoid the hard sample mining mechanism by using centers. 
Unlike normal contrastive center loss, our CMCC loss only constrains the distance among different centers and further optimizes feature distribution in the common feature space. 
It can be easily calculated and significantly improves the performance as a supplement to other losses. 

Suppose there are $P$ person identities, and each person has $K$ RGB images and $K$ IR images in a batch. 
Let $c^{m}_k$ represent the center of different modalities of person $k$, $k$ is the ID label of person, 
where $k \in \{1,2,...,n\}$, n is the number of person identities in a training batch. 
For each input $x_i^{m}$, we obtain feature $g_i^{m}$ by BN layer and $L_2$-norm after feature embedding. 
$c^m_k$ can be formulated as:
\begin{equation}
  c_k^{m} = \frac{1}{K} \sum\nolimits_{label(x_i^{m})=k} g_i^{m}
\end{equation}%

For each identity $k$ in a training batch, the feature center $c_k$ is the mean value of $c_k^{rgb}$ and $c_k^{ir}$ cause the number of RGB and IR images are the same. 
$d_{intra}$ is the Euclidean distance between the centers of different modalities of $k$ ($c_k^{rgb}$ and $c_k^{ir}$). 
$d_{inter}$ represents all the Euclidean distances between the $c_k$ and each feature center of other identities in the batch. 
As shown in Figure~\ref{Whole_Model}c, for any identity, $d_{intra}$ should be far less than each $d_{inter}$. 
The contrastive differences between $d_{inter}$ and $d_{intra}$ can be calculated as the distance of $d_{intra}$ and $\tilde{d}_{inter}$ ($\tilde{d}_{inter}$ represent the minimum of all $d_{inter}$). 
In other words, $d_{intra}$ should be far less than $\tilde{d}_{inter}$. 
In each batch, the $\mathcal{L}_{cmcc}$ can be formulated by: 
\begin{equation}
  \resizebox{0.8\linewidth}{!}{$
    \displaystyle
    \mathcal{L}_{cmcc} = \sum_{k=1}^n \frac{\log (1 + \exp (- (\tilde{d}_{inter} - d_{intra})))}{n}
    $}
    \label{cmcc}
\end{equation}%

By optimizing the $\mathcal{L}_{cmcc}$, we can decrease $d_{intra}$ while increase $\tilde{d}_{inter}$ simultaneously. 
Thus, the CMCC loss makes images from different identities more distinguishable while keeping features compact under the same identity.

\subsection{Cross-Modality Feature learning} \label{methodology_section_4}
In the modality-joint constraints, the Identity (ID) loss and weighted regularization triplet (WRT) loss are combined with CMCC loss to learn more discriminative modality-sharable feature. 
We calculate ID loss, WRT loss with the vectors of different positions as shown in Figure~\ref{Whole_Model}a. 
After Softmax operation, we get $p_i$ to represent the classification outputs of the image $x^m_i$, 
and each $x^m_i$ corresponds to an one-hot label list $y_i$. 
We define $p_i$ and $y_i$ as follows: 
\begin{equation}
  p_i=[p^1_i, p^2_i,...,p^j_i]
  \qquad
  y_i=[y^1_i, y^2_i,...,y^j_i]
\end{equation}%


Where $j \in \{1,2,...,N\}$ and N is the the number of person identities. 
ID loss ($\mathcal{L}_{id}$) can be expressed as
\begin{equation}
  \mathcal{L}_{id} = - \sum_{i=1}^B \sum_{j=1}^N y_i \log p^j_i 
  \label{id}
\end{equation}%


Following~\cite{BagOfTricks} and~\cite{AGW}, WRT loss ($\mathcal{L}_{wrt}$) is computed by the $L_2$-norm results of the feature vectors before BN layer as follows: 
\begin{equation}
  \resizebox{0.91\linewidth}{!}{$
    \displaystyle
    \mathcal{L}_{wrt}(x^m_i, x^m_{i+}, x^m_{i-}) = \log (1 + \exp ( w_i^p d_{i i+} - w_i^n d_{i i-}))
    $}
    \label{wrt}
\end{equation}%

\begin{equation}
  w_i^p = \frac{ \exp (d_{i i+}) }{ \sum_{d \in \mathcal{P}} \exp (d) },~~~ 
  w_i^n = \frac{ \exp (-d_{i i-}) }{ \sum_{d \in \mathcal{N}} \exp (-d) }
  \label{w}  
\end{equation}%

Where $(x^m_i, x^m_{i+}, x^m_{i-})$ represents a triple sample in training batch, 
containing one anchor sample $x^m_i$, one positive sample $x^m_{i+}$ with the same identity, and one negative sample $x^m_{i-}$ from a different identity. 
$w_i^p$ and $w_i^n$ are formulated by Equation~\ref{w} and $d$ is the Euclidean distance between the feature vectors. 
$\mathcal{P}$ is the set of all distances between every positive pair and $\mathcal{N}$ is the negative set. 

The total loss ($\mathcal{L}_{total}$) consists of PEF loss ($\mathcal{L}_{pef}$), ID loss ($\mathcal{L}_{id}$), WRT loss ($\mathcal{L}_{wrt}$), and CMCC loss ($\mathcal{L}_{cmcc}$): 
\begin{equation}
  \mathcal{L}_{total} = \mathcal{L}_{pef} + \mathcal{L}_{id} + \mathcal{L}_{wrt} + \mathcal{L}_{cmcc}
  \label{total}
\end{equation}%

$\mathcal{L}_{id}$ and $\mathcal{L}_{wrt}$ are calculated by Equation~\ref{id} and Equation~\ref{wrt}. 
The calculation methods of $\mathcal{L}_{pef}$ and $\mathcal{L}_{cmcc}$ are given in subsection~\ref{methodology_section_2} and subsection~\ref{methodology_section_3} respectively.

\section{Experiments}

\subsection{Datasets and Settings}


\paragraph{\textbf{Datasets:}} 
The proposed method is evaluated on the public dataset SYSU-MM01~\cite{Wu_2017_ICCV} and RegDB~\cite{RegDB}.
SYSU-MM01 is the largest RGB-IR ReID dataset used by mainstream methods on this task. 
This dataset contains 491 identities, consisting of 29,033 RGB images and 15,712 IR images taken from 4 RGB cameras and 2 IR cameras in indoor and outdoor environments. 
The training set contains 395 identities, including 22,258 RGB images and 11,909 IR images, while the test set has 96 identities with 3,803 IR images for the query set. 
As for the gallery, 301 or 3,010 (single-shot or multi-shot) RGB images are randomly selected to generate the gallery set. 
Following~\cite{Wu_2017_ICCV}, there are two evaluation modes for RGB-IR ReID: all-search with all images and indoor-search with only indoor images. 
RegDB dataset consists of 412 identities' images with 10 RGB images and 10 infrared for each identity, which is collected by a pair of aligned far-infrared and visible cameras.
According to the previous methods' partition strategy~\cite{AlignGAN,cmGAN,Sphere}, this dataset is divided equally into two halves for training and testing.
That is, the training set and test set each have 2,060 RGB and 2,060 infrared images.
During testing, it has two kinds of evaluation mode.
All RGB images/infrared images in the test set can be used as query, and all infrared images/RGB images can be used as gallery, corresponding to the Visible to Thermal/Thermal to Visible mode.

\begin{table*}[h]
  \caption{Comparison with state-of-the-art cross-modality ReID methods on the SYSU-MM01 dataset}
  \centering
  \fontsize{9}{11.55}\selectfont
  \setlength{\tabcolsep}{1.9mm}
  \begin{threeparttable}
  \begin{tabular}{l|c|c c c|c c c|c c c|c c c}
  \toprule[0.5pt]
  \hline
  \multirow{3}{*}{Methods} & \multirow{3}{*}{Venue} 
  & \multicolumn{6}{c}{\textit{All-Search}} & \multicolumn{6}{|c}{\textit{Indoor-Search}} \cr
  \cline{3-14}
  & ~ & \multicolumn{3}{c}{\textit{Single-Shot}} & \multicolumn{3}{|c}{\textit{Multi-Shot}} & \multicolumn{3}{|c}{\textit{Single-Shot}} & \multicolumn{3}{|c}{\textit{Multi-Shot}} \cr
  & ~ & R1 & R10 & mAP & R1 & R10 & mAP & R1 & R10 & mAP & R1 & R10 & mAP \cr
    \hline
    Two-Stream\cite{Wu_2017_ICCV}   & ICCV 17 & 11.65 & 47.99 & 12.85 & 16.33 & 58.35 & 8.03  & 15.60 & 61.18 & 21.49 & 22.49 & 72.22 & 13.92 \cr
    One-Stream\cite{Wu_2017_ICCV}   & ICCV 17 & 12.04 & 49.68 & 13.67 & 16.26 & 58.14 & 8.59  & 16.94 & 63.55 & 22.95 & 22.62 & 71.74 & 15.04 \cr
    Zero-Padding\cite{Wu_2017_ICCV} & ICCV 17 & 14.80 & 54.12 & 15.95 & 19.13 & 61.40 & 10.89 & 20.58 & 68.38 & 26.92 & 24.43 & 75.86 & 18.64 \cr
    BDTR\cite{dc}                   & IJCAI 18 & 17.01 & 55.43 & 19.66 & -     & -     & -     & -     & -     & -     & -     & -     & -    \cr
    HSME\cite{Sphere}               & AAAI 19 & 18.03 & 58.31 & 19.98 & -     & -     & -     & -     & -     & -     & -     & -     & -     \cr
    D-HSME\cite{Sphere}             & AAAI 19 & 20.68 & 62.74 & 23.12 & -     & -     & -     & -     & -     & -     & -     & -     & -    \cr
    SDL\cite{SDL}                   & TCSVT 20 & 28.12 & 70.23 & 29.01 & -     & -     & -     & 32.56 & 80.45 & 39.56 & -     & -     & -     \cr
    DGD+MSR\cite{DGD}               & TIP 19 & 37.35 & 83.40 & 38.11 & 43.86 & 86.94 & 30.48 & 39.64 & 89.29 & 50.88 & 46.56 & 93.57 & 40.08 \cr
    EDFL\cite{EDFL}                 & NeuroC 20 & 36.94 & 84.52 & 40.77 & -     & -     & -     & -     & -     & -     & -     & -     & -     \cr
    HPILN\cite{HP}                  & IET-IPR 19 & 41.36 & 84.78 & 42.95 & 47.56 & 88.13 & 36.08 & 45.77 & 91.82 & 56.52 & 53.05 & 93.71 & 47.48 \cr
    AGW\cite{AGW}                   & TPAMI 21 & 47.50 & -     & 47.65 & 50.87 & -     & 40.03 & 54.17 & -     & 62.97 & -     & -     & -     \cr
    TSLFN+HC\cite{HC}               & NeuroC 20 & 56.96 & 91.50 & 54.95 & 62.09 & 93.74 & 48.02 & 59.74 & 92.07 & 64.91 & 69.76 & 95.85 & 57.81 \cr
    \hline
    cmGAN\cite{cmGAN}               & IJCAI18 & 26.97 & 67.51 & 27.80 & 31.49 & 72.74 & 22.27 & 31.63 & 77.23 & 42.19 & 37.00 & 80.94 & 32.76 \cr
    $D^2$RL\cite{DRL}               & CVPR 19 & 28.90 & 70.60 & 29.20 & -     & -     & -     & -     & -     & -     & -     & -     & -     \cr
    Hi-CMD\cite{Hi-CMD}             & CVPR 20 & 34.94 & 77.58 & 35.94 & -     & -     & -     & -     & -     & -     & -     & -     & -     \cr
    JSIA\cite{JSIA}                 & AAAI 20 & 38.10 & 80.70 & 36.90 & 45.10 & 85.70 & 29.50 & 43.80 & 86.20 & 52.90 & 52.70 & 91.10 & 42.70 \cr
    AlignGAN\cite{AlignGAN}         & ICCV 19 & 42.40 & 85.00 & 40.70 & 51.50 & 89.40 & 33.90 & 45.90 & 87.60 & 54.30 & 57.10 & 92.70 & 45.30 \cr
    tsGAN\cite{tsGAN}               & arXiv 20 & 49.80 & 87.30 & 47.40 & 56.10  & 90.20 & 38.50 & 50.40 & 90.80 & 63.10 & 59.30 & 91.20 & 50.20 \cr
    \hline
    X-Modality\cite{X}              & AAAI 20 & 49.92 & 89.79 & 50.73 & -     & -     & -     & -     & -     & -     & -     & -     & -     \cr
    CMM+CML\cite{CMM+CML}           & ACMMM 20 & 51.80 & \textbf{92.72} & 51.21 & 56.27 & 94.08 & 43.39 & 54.98 & 94.38 & 63.70 & 60.42 & 96.88 & 53.52  \cr
    HAT\cite{HAT}                   & TIFS 20 & 55.29 & 92.14 & 53.89 & -     & -     & -     & 62.10 & 95.75 & 69.37 & -     & -     & -     \cr

    \hline
    MSO (Ours)& - & \textbf{58.70} & 92.06 & \textbf{56.42} & \textbf{65.85} & \textbf{94.37} & \textbf{49.56} & \textbf{63.09} & \textbf{96.61} & \textbf{70.31} & \textbf{72.06} & \textbf{97.77} & \textbf{61.69} \cr
    \hline
    \bottomrule[0.5pt]
  \end{tabular}
  \end{threeparttable}
  \label{comparison}
  \end{table*}

  \paragraph{\textbf{Evaluation metrics:}}
  Our experiments use the standard Cumulative Matching Characteristics (CMC) curve and the mean average precision (mAP) as the evaluation metrics. 
  CMC is represented as rank-k, which tells the rate where the correct match is within the k-nearest neighbors, with k in our results are 1, 10, 20. 
  In the ablation experiments, following~\cite{AGW}, we also compute the mean inverse negative penalty (mINP) as one of the evaluation metrics. 
  When testing on the SYSU-MM01 dataset, we evaluate methods on the specified test set and the randomly generated gallery, repeat 10 times, then take the average value to measure models' performance~\cite{Wu_2017_ICCV}. 
  As for the evaluation of the RegDB dataset, we adopt its evaluation protocol described above, doing experiments in both Visible to Thermal and Thermal to Visible mode to fully verify the effectiveness of our proposed method.

\paragraph{\textbf{Implementation details:}}
We implement the proposed model based on the deep learning framework PyTorch. 
In Section~\ref{methodology_section_1}, we have explained the structure of the model in detail. 
The backbone network, ResNet-50, is initialized with the parameters pretrained on the ImageNet. 
Following~\cite{AGW}, the stride of the last convolutional layer in ResNet-50 is changed to 1. 
All the input images are uniformly resized to $288 \times 144$. 
During the training phase, we use several simple data augmentation strategies that include random crop and horizontal flip. 
Each mini-batch contains 64 images of 8 identities, which means 4 pairs of RGB and infrared images are selected randomly from each identity. 
The whole model is optimized with Adam for 100 epochs with an initial learning rate of 0.0005, and we decay the learning rate by 0.1 at epoch 20, 25, and 35, respectively. 
When testing the method, we use cosine distance to measure the difference between query and gallery's extracted features, generating retrieval results of this ReID task.

\begin{table}[t]
  \caption{Comparison with state-of-the-art cross-modality ReID methods on the RegDB dataset}
  \centering
  \fontsize{7}{11.3}\selectfont
  \setlength{\tabcolsep}{1.9mm}
  \begin{tabular}{l|c|c c c|c c c}
    \hline
    \hline
    \multirow{2}{*}{Methods} & \multirow{2}{*}{Venue}
    & \multicolumn{3}{c}{\textit{Visible to Thermal}} & \multicolumn{3}{|c}{\textit{Thermal to Visible}} \cr
    & ~ & R1 & R10 & mAP & R1 & R10 & mAP \cr
    \hline
    Zero-Padding\cite{Wu_2017_ICCV} & ICCV 17 & 17.8 & 34.2  & 18.9 & 16.6 & 34.7  & 17.8 \cr 
    BDTR\cite{dc}                   & IJCAI 18 & 33.5 & 58.4  & 31.8 & 32.7 & 58.0  & 31.1 \cr
    HSME\cite{Sphere}               & AAAI 19 & 41.3 & 65.2  & 38.8 & 40.7 & 65.4  & 37.5 \cr
    D-HSME\cite{Sphere}             & AAAI 19 & 50.9 & 73.4 & 47.0 & 50.2 & 72.4  & 46.2 \cr
    SDL\cite{SDL}                   & TCSVT 20 & 26.5 & 51.3 & 23.6 & 25.7 & 50.2  & 22.9 \cr
    DGD+MSR\cite{DGD}               & TIP 19 & 48.4 & 70.3 & 48.7 & -    & -    & -    \cr
    EDFL\cite{EDFL}                 & NeuroC 20 & 52.6 & 72.1 & 53.0 & -    & -   & -    \cr
    AGW\cite{AGW}                   & TPAMI 21 & 70.0 & -  & 66.4 & -    & -    & -    \cr
    cmGAN\cite{cmGAN}               & IJCAI18 & 47.9 & 47.9 & 12.8 & 16.3 & 47.9 & 65.5 \cr
    $D^2$RL\cite{DRL}               & CVPR 19 & 43.4 & 66.1 & 44.1 & -    & -    & -    \cr
    Hi-CMD\cite{Hi-CMD}             & CVPR 20 & 70.9 & 86.4  & 66.0 & -    & -    & -    \cr
    JSIA\cite{JSIA}                 & AAAI 20 & 48.5 & -     & 49.3 & 48.1 & -    & 48.9 \cr
    AlignGAN\cite{AlignGAN}         & ICCV 19 & 57.9 & -    & 53.6 & 56.3 & -    & 53.4 \cr
    cm-SSFT\cite{ssft}              & CVPR 20 & 65.4 & -    & 65.6 & 63.8 & -    & 64.2 \cr
    X-Modality\cite{X}              & AAAI 20 & -    & -    & -    & 62.2 & 83.1 & 60.2 \cr
    CMM+CML\cite{CMM+CML}           & ACMMM 20 & -    & -     & -    & 59.8 & 80.4 & 60.9 \cr
    HAT\cite{HAT}                   & TIFS 20 & 71.8 & 87.2 & \textbf{67.6} & 70.0 & 86.5 & 66.3 \cr
    \hline
    MSO (Ours)                      & - & \textbf{73.6} & \textbf{88.6} & 66.9 & \textbf{74.6} & \textbf{88.7} & \textbf{67.5} \cr
    \hline
    \hline
  \end{tabular}
  \label{comparison_regdb}
\end{table}

\begin{figure*}[t]
  \subfloat[]{\includegraphics[width=0.48\linewidth]{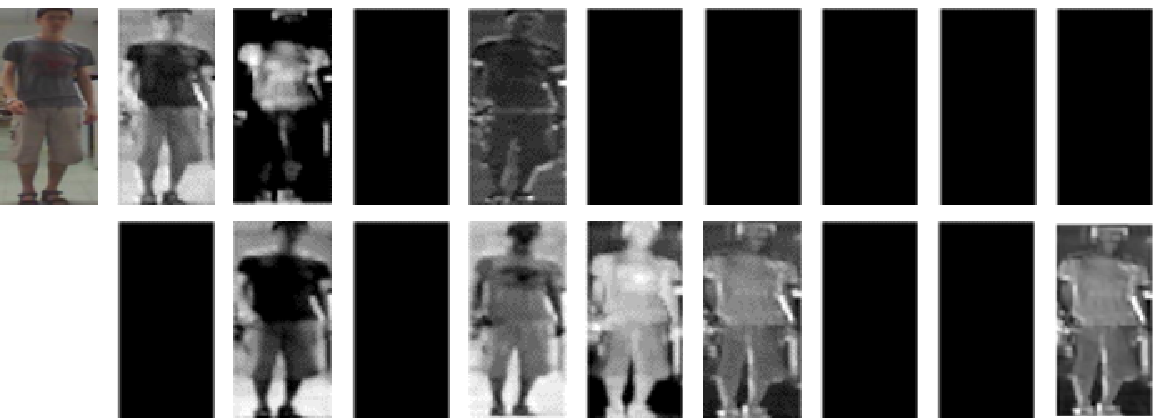} \label{vfa}}
  \hspace{0.01\linewidth}
  \subfloat[]{\includegraphics[width=0.48\linewidth]{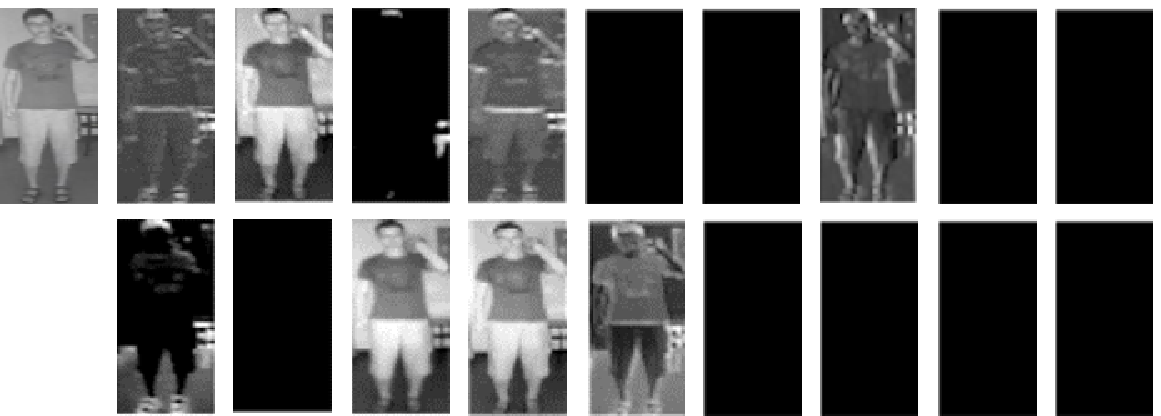} \label{vfb}}
  \vspace{0.01\linewidth}
  \subfloat[]{\includegraphics[width=0.48\linewidth]{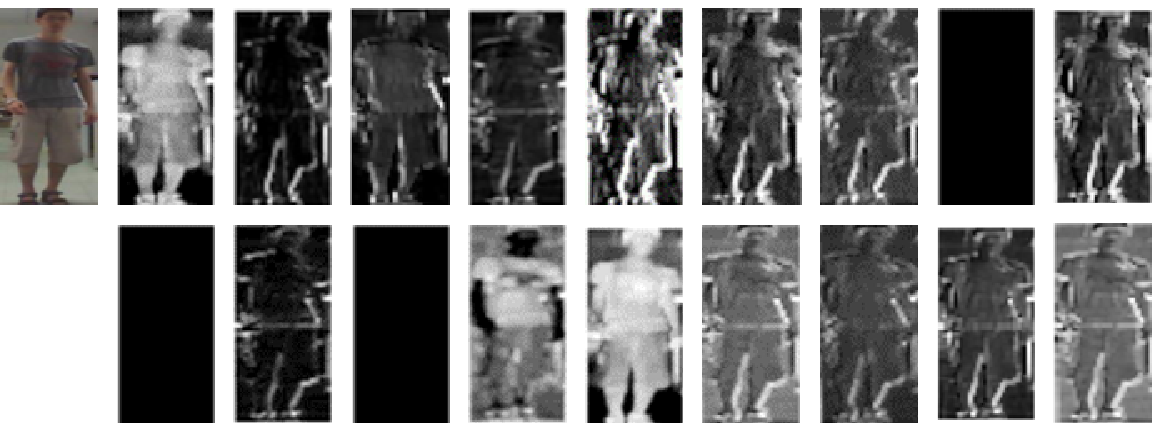} \label{vfc}}
  \hspace{0.01\linewidth}
  \subfloat[]{\includegraphics[width=0.48\linewidth]{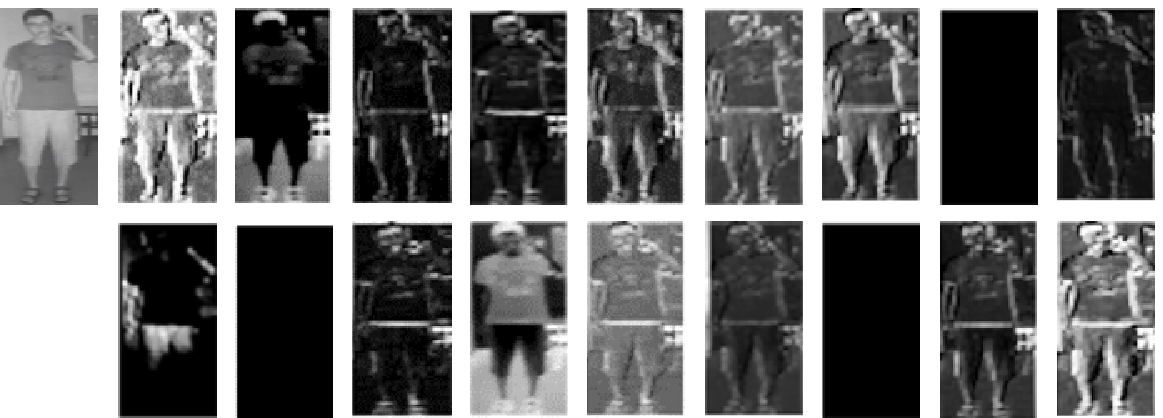} \label{vfd}}
  \caption{feature maps visualization. 
  (a) Results of baseline with RGB image as input. 
  (b) Results of baseline with IR image of the same identity as input. 
  (c) Results of baseline+PEF with the same RGB image as (a). 
  (d) Results of baseline+PEF with the same IR image as (b).}
  \label{visiblefeature}
  \vspace{-0.02\linewidth}
\end{figure*}

\subsection{Comparison with State-of-the-art Methods}

In this subsection, we compare our proposed method with several cross-modality person ReID methods that include the following categories: 
1) With different structures and loss functions, Two-Stream, One-Stream, Zero-Padding~\cite{Wu_2017_ICCV}, HSME, D-HSME~\cite{Sphere}, BDTR, SDL~\cite{SDL}, DGD+MSR~\cite{DGD}, EDFL~\cite{EDFL}, HPILN~\cite{HP}, AGW~\cite{AGW}, cm-SSFT~\cite{ssft}, and TSLFN+HC~\cite{HC} learned modality-invariant feature representation; 
2) With the ideas of GAN, cmGAN~\cite{cmGAN}, $D^2$RL~\cite{DRL}, Hi-CMD~\cite{Hi-CMD}, JSIA~\cite{JSIA}, AlignGAN~\cite{AlignGAN}, and tsGAN~\cite{tsGAN} generated cross-modality images or features; 
3) X-Modality~\cite{X}, CMM+CML~\cite{CMM+CML}, and HAT~\cite{HAT} introduced a third modality to feature space as a three-modality learning problem. 

The experimental results of these state-of-the-art methods on two datasets SYSU-MM01 and RegDB are shown in Table~\ref{comparison} and~\ref{comparison_regdb}. 
The R1, R10, R20 denote Rank-1, Rank-10, and Rank-20 accuracies (\%), respectively. 
And the mAP denotes the mean average precision score (\%). 
As shown in Table~\ref{comparison}, our proposed model shows great performance. 
Compared with TSLFN+HC~\cite{HC}, which achieved optimal performance with local features, our model over 1.74\% on rank-1 and 1.47\% on mAP using only global features in the single-shot setting of all-search mode. 
In the single-shot setting of indoor-search mode, our model achieves a rank-1 accuracy of 63.09\% and an mAP of 70.31\%, which are higher than TSLFN+HC by 3.35\% and 5.40\%, respectively. 
Compared with other methods that also only use global features, the proposed method outperforms other methods by a large margin. 
Compared with the results in the single-shot setting of all-search mode, 
our model is 11.20\% and 8.77\% higher than the strong AGW model on Rank-1 and mAP respectively, which has the same backbone. 

As shown in Table~\ref{comparison_regdb}, our method also achieves much higher accuracy on the evaluation of the RegDB dataset. 
In the Visible to Thermal mode, the proposed model achieves a rank-1 accuracy of 73.6\% and a rank-10 accuracy of 88.6\%, which are higher than other methods. 
Compared with the latest HAT model~\cite{HAT}, our method also achieves similar performance on mAP. 
In the Thermal to Visible mode, the proposed model surpasses the method HAT by 4.6\% on Rank-1 and 1.2\% on mAP, surpasses the method CMM+CML~\cite{CMM+CML} by 14.8\% on Rank-1 and 6.6\% on mAP.

\subsection{Ablation Study}
In this subsection, we design the ablation experiments to test the effectiveness of PEF loss and CMCC loss. 
All ablation experiments are performed on the dataset SYSU-MM01, using the single-shot setting of all-search mode (the more difficult mode). 
Specifically, "B" represents the baseline model using the proposed network without PEF loss and CMCC loss. 

\begin{table}[t]
  \caption{Ablation studies on the SYSU-MM01 dataset}
  \centering
  \fontsize{9}{15}\selectfont
  \setlength{\tabcolsep}{1.9mm}
  \begin{tabular}{l|c c c c c}
    \hline
    \hline
    Methods & R1 & R10 & R20 & mAP & mINP \cr
    \hline
    Baseline (B)      & 49.16 & 86.02 & 93.25 & 46.98 & 32.75 \cr
    B+PEF             & 52.58 & 87.92 & 94.69 & 49.98 & 35.44 \cr
    B+expAT\cite{Triplet0}  & 52.01 & 89.16 & 95.40 & 49.78 & 35.36 \cr
    B+TC\cite{Triplet2}     & 53.37 & 89.73 & 95.68 & 51.47 & 37.49 \cr
    B+HC\cite{HC}  ($\lambda$ = 0.1) & 53.84 & 88.89 & 94.87 & 48.68 & 31.13 \cr
    B+CMCC            & 56.63 & 91.83 & 96.90 & 54.93 & 40.85  \cr
    B+PEF+CMCC        & \textbf{58.70} & \textbf{92.06} & \textbf{97.20} & \textbf{56.42} & \textbf{42.04} \cr
    \hline
    \hline
  \end{tabular}
  \label{ablation}
\end{table}

\begin{figure*}[t]
  \centering
  \subfloat[]{\includegraphics[height=0.22\linewidth]{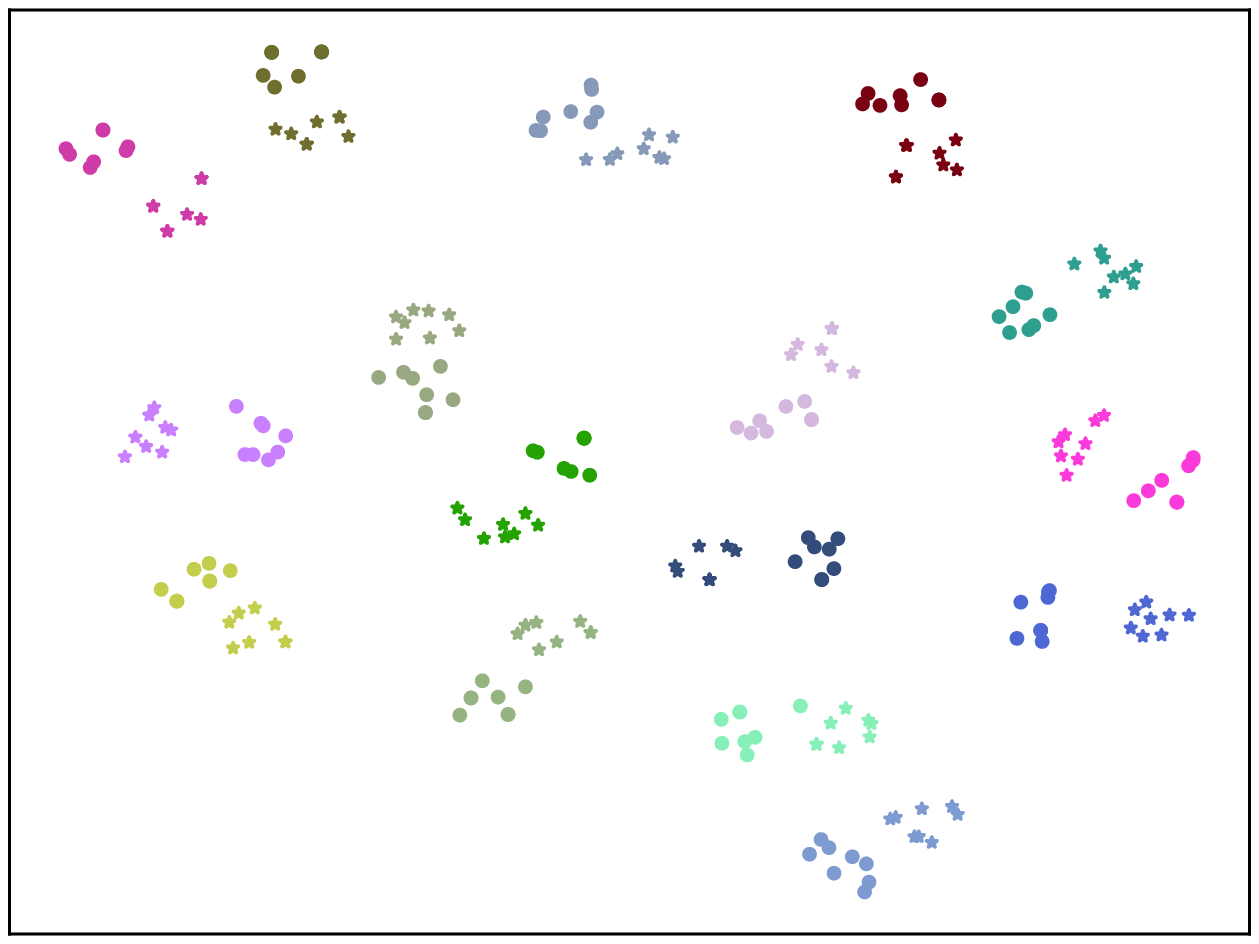} \label{baseline}}
  \hspace{0.00005\linewidth}
  \subfloat[]{\includegraphics[height=0.22\linewidth]{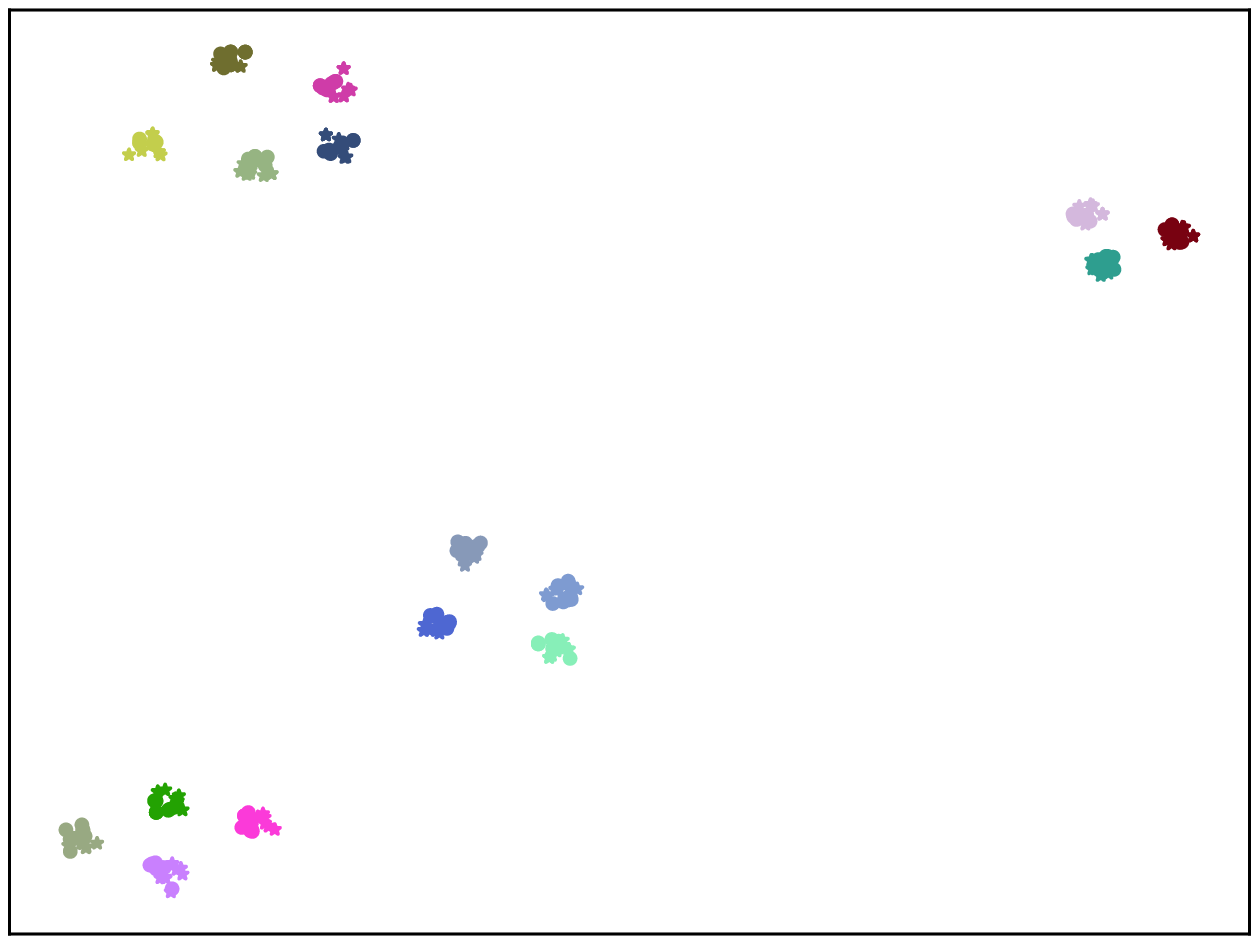} \label{baseline_hc}}
  \hspace{0.00005\linewidth}
  \subfloat[]{\includegraphics[height=0.22\linewidth]{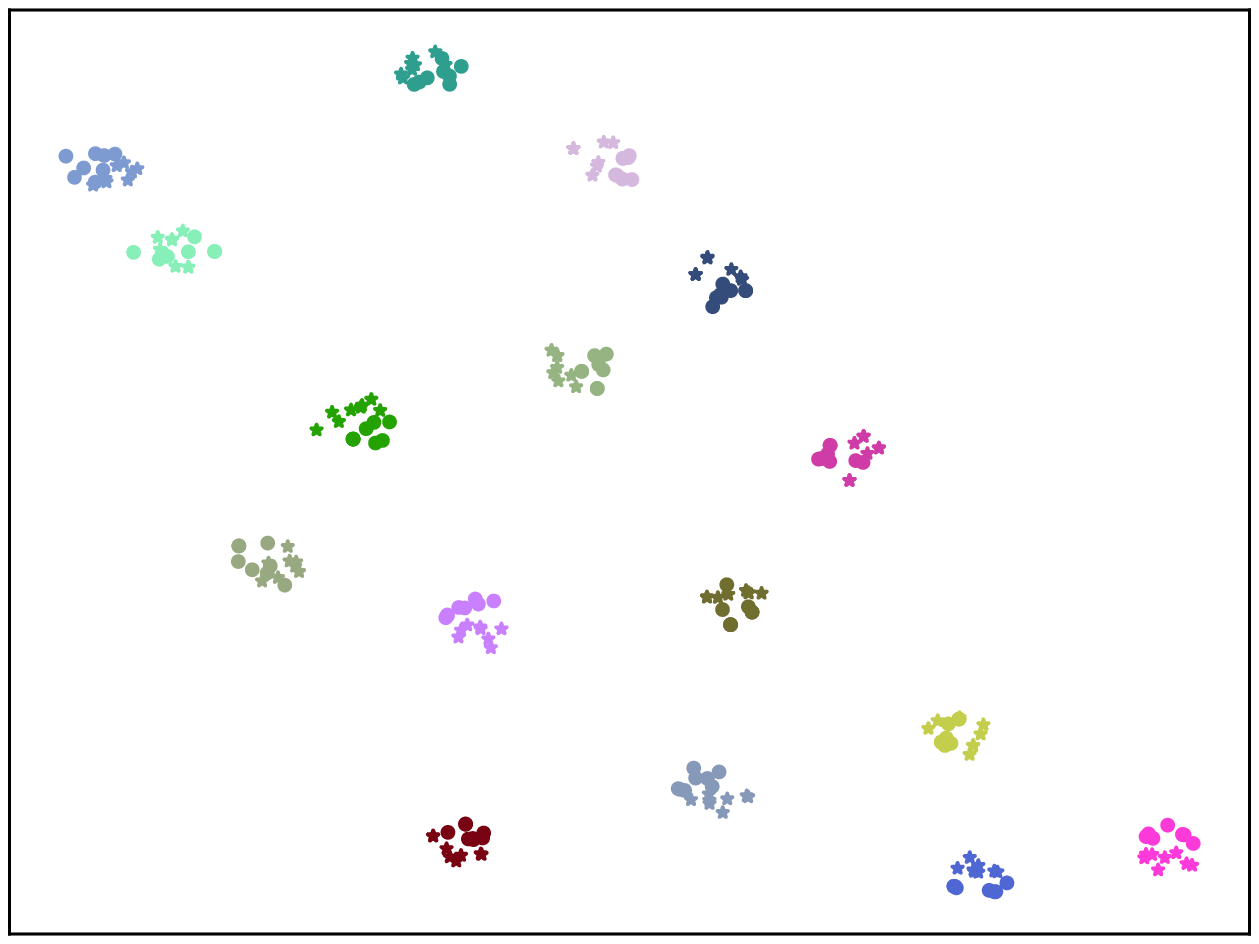} \label{baseline_cmcc}}
  \hspace{0.00003\linewidth}
  \subfloat[]{\includegraphics[height=0.219\linewidth]{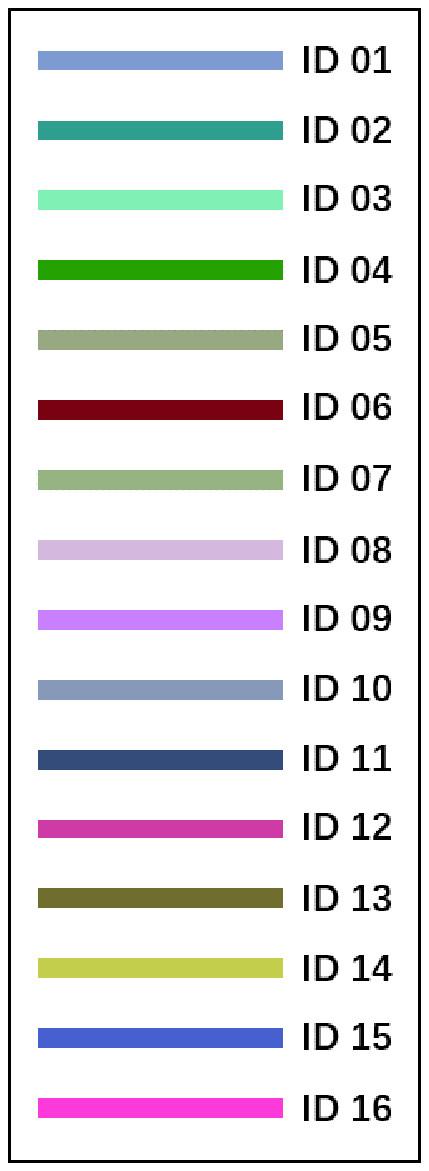} \label{ids}}
  \caption{Feature distributions visualized with t-SNE. 
  (a) Feature distribution of baseline; 
  (b) Feature distribution of baseline+HC; 
  (c) Feature distribution of baseline+CMCC.}
  \label{feature_distribution}
\end{figure*}

\paragraph{\textbf{Influence of PEF Loss:}}
The results of ablation experiments for PEF loss are shown in Table~\ref{ablation}. 
Compared with the baseline model (B), the rank-1 accuracy and mAP gains 3.42\% and 3.00\% improvements respectively by introducing PEF loss (B+PEF), 
which proves the effectiveness of PEF loss. 
Additionally, we respectively visualize the shallow feature maps of the B model and the B+PEF model. 
Specifically, we take one RGB image and one IR image of the same person identity as input and extract their shallow feature maps of these models. 
As shown in Figure~\ref{visiblefeature}, we visualize RGB feature maps and IR feature maps extracted from the selected images. 
Figure~\ref{visiblefeature}a and Figure~\ref{visiblefeature}b show the feature maps of the B model, and Figure~\ref{visiblefeature}c and Figure~\ref{visiblefeature}d show the feature maps of the B+PEF model.
The black images of feature maps represent inactive feature maps with zero values, which are useless for modality-sharable features learning. 
As shown in Figure~\ref{visiblefeature}, no matter which modality the input image belongs to, more useful non-zero feature maps are extracted after introducing the PEF loss, 
and the edge information in the feature maps are enhanced obviously. 
Besides, mAP decreases after removing PEF loss from the overall model MSO (B+PEF+CMCC), which also demonstrates the effectiveness of PEF loss.

\paragraph{\textbf{Influence of CMCC Loss:}}
As shown in Table~\ref{ablation}, the model with CMCC loss (B+CMCC) achieves a rank-1 accuracy of 56.63\% and an mAP of 54.93\%, which are higher than the baseline (B) by 7.47\% and 7.95\%, respectively. 
Besides, we implement expAT loss~\cite{Triplet0} and triplet center loss (TC)~\cite{Triplet2} with the baseline respectively. 
Compared with B+expAT and B+TC, the addition of CMCC loss brings a marked performance boost to baseline. 
We also introduce hetero-center loss (HC)~\cite{HC} into the baseline (B+HC).
The results of B+HC in Table~\ref{ablation} are the highest results of HC in several experiments with different values of $\lambda$~\cite{HC}. 
The advantage of CMCC loss is evident because CMCC loss outperforms HC loss by 2.79\% in rank-1 accuracy and 6.25\% in mAP.  
In addition, we visualize the distribution of features learned by baseline model, baseline with HC loss model, and baseline with our CMCC loss model respectively. 
As shown in Figure~\ref{feature_distribution}, the feature distribution of baseline shows low discrimination, 
and the distribution of B+HC has small margin of different modalities of the same identity but large similarity of different identity. 
Compared with B and B+HC, our CMCC loss is effective on separating different identities as shown in Figure~\ref{baseline_cmcc}.

\subsection{Edge Fusion Strategy Analysis} \label{Analysis}

In this subsection, we try four other different edge fusion strategies as the edge features enhancement module. 
Among them, three edge fusion strategies is used to enhance the modality-sharable features in each single-modality space: 
1) \textbf{Directly Add Fusion:} Get the feature maps after layer0 and the edge features extracted by sobel convolution module, directly add edge features to each feature map. 
2) \textbf{Weighted Add Fusion:} Add edge features to each feature map with different learnable weights. 
3) \textbf{Concat Fusion:} Get the feature maps of 64 channels after layer0, and concatenate them with the edge features of 1 channel. 
Then change back to 64 channels through a 1×1 convolutional layer. 
The experimental results with single-shot setting of all-search mode on SYSU-MM01 are shown in Table~\ref{comparison_fusion_strategy}. 
Compared with above three edge fusion strategies, PEF loss shows best performance in enhancing modality-sharable feature learning on cross-modality person ReID task. 

To prove the significant advantages of placing the edge features enhancement module in each single-modality space, 
we also design another method: 4) \textbf{Classic Feature Fusion:} Take the edge features extracted from sobel convolution module as the third modality and feed them into convolutional neural network with RGB and IR images. 
Then concatenate image features with their edge features after CNN and use FC layers to fuse them. 
This means that the edge information are enhanced in the common feature space. 
As shown in Table~\ref{comparison_fusion_strategy}, the performance of this classic feature fusion method is relatively poor, 
which proves the rationality and effectiveness of optimization in single-modality space.

\begin{table}[t]
  \caption{Comparison of different edge fusion strategies}
  \centering
  \fontsize{8.5}{12}\selectfont
  \setlength{\tabcolsep}{1.9mm}
  \begin{tabular}{l|c c c c c}
    \hline
    \hline
    Strategies & R1 & R10 & R20 & mAP & mINP \cr
    \hline
    Directly Add Fusion & 55.56 & 90.00 & 95.71 & 53.67 & 40.03 \cr
    Weighted Add Fusion & 56.77 & 91.63 & 96.88 & 54.47 & 40.06 \cr
    Concat Fusion       & 56.33 & 92.28 & 97.18 & 54.19 & 40.00 \cr
    PEF Loss Fusion     & \textbf{58.70} & \textbf{92.06} & \textbf{97.20} & \textbf{56.42} & \textbf{42.04} \cr
    \hline
    Classic Feature Fusion     & 50.33 & 88.04 & 94.82 & 49.51 & 35.84 \cr
    \hline
    \hline
  \end{tabular}
  \label{comparison_fusion_strategy}
\end{table}

\section{Conclusion}
In this paper, we present an innovative method for RGB-IR cross-modality person ReID, the multi-feature space joint optimization (MSO) network, which can learn modality-sharable features in both the single-modality space and the common space.
Based on the observation that edge information is modality-invariant, we propose an edge features enhancement module to enhance the modality-sharable features in each single-modality space. 
In our method, the perceptual edge features (PEF) loss is introduced in the feature space of each modality, and cross-modality contrastive-center (CMCC) loss is introduced in the common feature space, 
which can enhance the modality-sharable features and learn more discriminative feature distribution. 
Through edge fusion strategy analysis, we also prove that PEF loss has outstanding advantages over the other fusion strategies. 
Experiments show that the performance of our proposed method is significantly improved against the state-of-the-art methods on both the SYSU-MM01 and RegDB datasets. 
Code will be made available. 
We believe that the new method will provide innovative solutions for future cross-modality ReID research.








\begin{acks}
This work was supported by the National Natural Science Foundation of China (Nos.61972030), 
and the Grapevine Scholar Plan of JD AI Research.
\end{acks}

\bibliographystyle{ACM-Reference-Format}
\balance
\bibliography{sample-xelatex}










\end{document}